\documentclass[letterpaper, 10 pt, conference]{ieeeconf}
\makeatletter
\let\NAT@parse\undefined
\makeatother
\IEEEoverridecommandlockouts                              %

\usepackage[numbers,sort&compress]{natbib}

\usepackage{multirow}
\usepackage{multicol}
\usepackage[bookmarks=true]{hyperref}
\usepackage[dvipsnames]{xcolor}
\usepackage{tablefootnote} %
\usepackage{caption}
\usepackage{subcaption}
\usepackage{graphicx}
\usepackage{float}
\usepackage{amssymb}
\usepackage{amsmath}

\DeclareMathOperator*{\argmin}{argmin}
\usepackage{algorithm}
\usepackage{algpseudocode}
\usepackage{lipsum} 
\usepackage{tabularx}

\usepackage{sidecap}
\usepackage{hyphenat}
\usepackage{makecell}

\usepackage{xspace}
\def\eg{\textit{e.g.}\@\xspace}
\def\ie{\textit{i.e.}\@\xspace}

\newcommand{\R}{\mathbb{R}}

\begin{document}
\bstctlcite{BSTcontrol}

\title{\LARGE \bf
ViTa-Zero: Zero-shot Visuotactile Object 6D Pose Estimation
}

\author{Hongyu Li$^{1,2\dagger}$, James Akl$^1$, Srinath Sridhar$^{1,2}$, Tye Brady$^1$, Ta\c{s}k{\i}n Pad{\i}r$^{1,3\ddagger}$%
\thanks{$^\dagger$The work was done when Hongyu Li was an intern at Amazon. Corresponding to: \tt\small hli230@cs.brown.edu}
\thanks{
$^\ddagger$Ta\c{s}k{\i}n Pad{\i}r holds concurrent appointments as a Professor of Electrical and Computer Engineering at Northeastern University and as an Amazon Scholar. This paper describes work performed at Amazon and is not associated with Northeastern University.
}
\thanks{
$^{1}$Amazon Fulfillment Technologies \& Robotics, Westborough, MA, 01581. 
}%
\thanks{$^{2}$Brown University, Providence, RI, 02912.}%
\thanks{$^{3}$Northeastern University, Boston, MA, 02120.}%
}

\maketitle

\begin{abstract}
Object 6D pose estimation is a critical challenge in robotics, particularly for manipulation tasks. While prior research combining visual and tactile (visuotactile) information has shown promise, these approaches often struggle with generalization due to the limited availability of visuotactile data. In this paper, we introduce \textit{ViTa-Zero}, a zero-shot visuotactile pose estimation framework. Our key innovation lies in leveraging a visual model as its backbone and performing feasibility checking and test-time optimization based on physical constraints derived from tactile and proprioceptive observations. Specifically, we model the gripper-object interaction as a spring–mass system, where tactile sensors induce attractive forces, and proprioception generates repulsive forces. We validate our framework through experiments on a real-world robot setup, demonstrating its effectiveness across representative visual backbones and manipulation scenarios, including grasping, object picking, and bimanual handover. Compared to the visual models, our approach overcomes some drastic failure modes while tracking the in-hand object pose. In our experiments, our approach shows an average increase of 55\% in AUC of ADD-S and 60\% in ADD, along with an 80\% lower position error compared to FoundationPose.

\end{abstract}

\IEEEpeerreviewmaketitle

\section{Introduction}
For robots to intelligently manipulate objects like humans, they must perceive the object's state within their environment accurately.
A key aspect of this perception is object 6D pose estimation, \ie, estimating position and orientation in 3D space, which serves as a critical representation of object state~\cite{kroemer_review_2021}.
Accurate pose estimation can improve the performance of state-based manipulation policies and lead to improved learning efficiency and success rate~\cite{chen_visual_2023, chen_system_2022, qi_general_2023, thalhammer_challenges_2024, openai_solving_2019, andrychowicz_learning_2020, handa_dextreme_2023}.
Prior research has focused on instance-level~\cite{xiang_posecnn_2018, wang_densefusion_2019, park_pix2pose_2019, li_mrc-net_2024}, category-level~\cite{wang_normalized_2019, chen_learning_20200, lee_tta-cope_2023} and more recently novel object estimation~\cite{wen_foundationpose_2024, labbe_megapose_2022, he_onepose_2022, liu_gen6d_2022, lin_sam-6d_2024}.
Despite these advancements, state-based approaches still face significant challenges in practical, real-world applications compared to end-to-end visuomotor approaches~\cite{chen_visual_2023, chen_system_2022, chi_diffusion_2023, ze_3d_2024}.
This is due to the challenges of pose estimation in real-world scenarios, particularly during contact-rich and in-hand manipulation tasks characterized by frequent occlusions and dynamic interactions~\cite{thalhammer_challenges_2024, suresh_neuralfeels_2024}.

\begin{figure}[!ht]
    \centering
    \includegraphics[width=.9\linewidth]{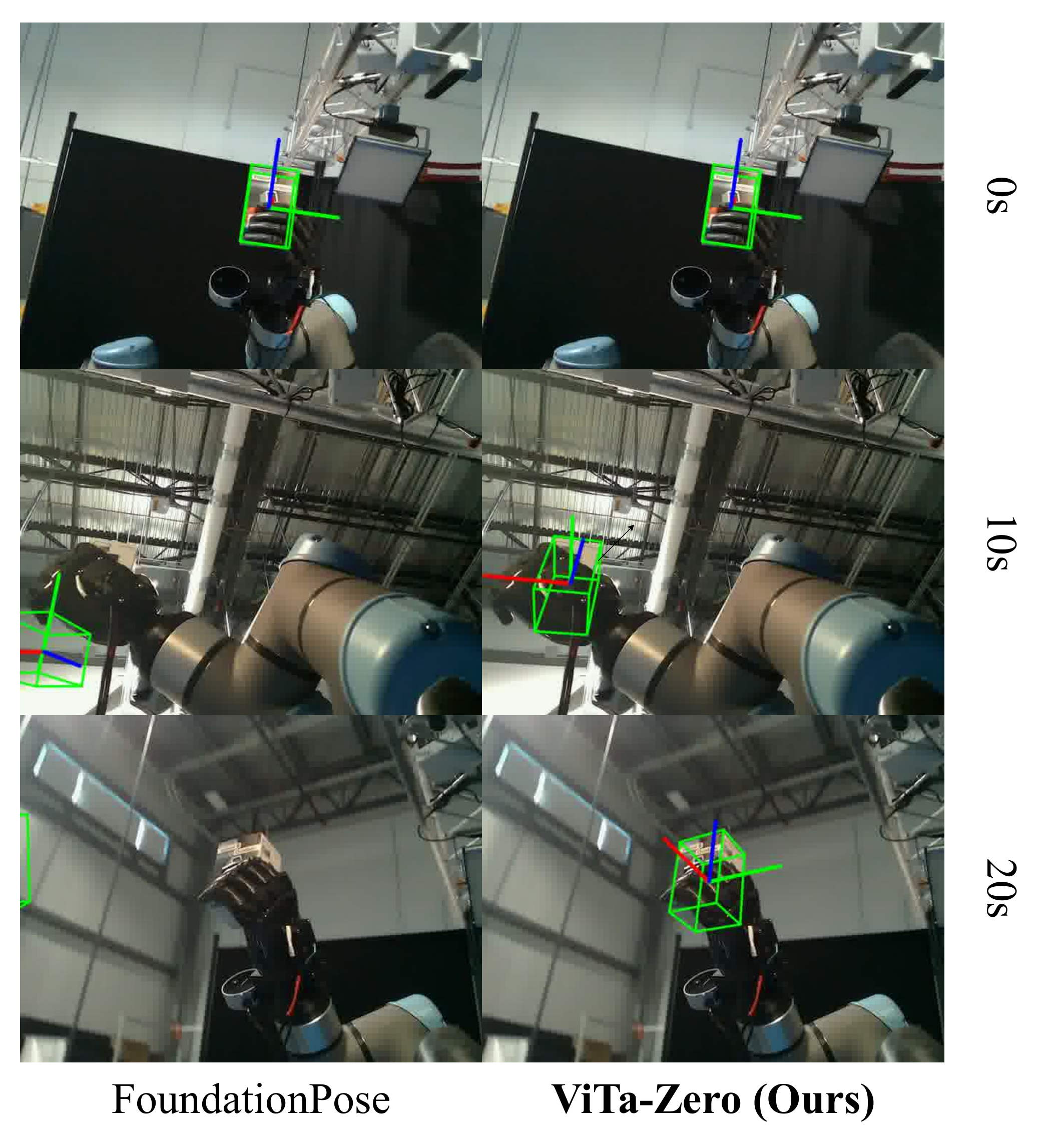}
    \caption{
    FoundationPose~\cite{wen_foundationpose_2024} (left) fails due to errors that are acute (\eg, occlusions) or accumulative (\eg, noise) while tracking the in-hand object. 
    Our approach (right) leverages tactile and proprioceptive observations for stable tracking.
    }
    \label{fig:teaser}
\end{figure}

To better estimate the pose of in-hand objects, previous works have explored the combination of visual and tactile (visuotactile) sensory observations~\cite{li_vihope_2023, suresh_neuralfeels_2024, dikhale_visuotactile_2022, wan_vint-6d_2024, rezazadeh_hierarchical_2023, tu_posefusion_2023, liu_enhancing_2024, gao_-hand_2023, li_hypertaxel_2024, li_v-hop_2025}. 
These works have shown improvements in grasping and in-hand manipulation tasks, as tactile sensors can reveal occluded parts of objects to enhance pose estimation during manipulation.
However, this introduces some practical challenges:
(1) collecting a visuotactile dataset is arduous due to the fragility and diversity of tactile sensors, making it hard to scale in the real world~\cite{yang_touch_2022}. 
While simulation offers a potential solution, the sim-to-real gap remains substantial~\cite{dikhale_visuotactile_2022, wan_vint-6d_2024, wang_tacto_2022, andrychowicz_learning_2020};
(2) visuotactile models often overfit to a single hardware setup and scenarios involving rich contact signals, such as in-hand object pose estimation, and struggle to perform on par with the visual counterparts for static and unoccluded scenes; and,
(3) adopting the generalization techniques used for visual models for these visuotactile models is non-trivial due to the inclusion of tactile data and the non-uniformity of tactile sensors and grippers.
Therefore, most models are sensor-specific and instance-level, limiting their practicality.

In this paper, we address these challenges by proposing \textbf{ViTa-Zero}, a framework that estimates and tracks the 6D pose of the novel object through the \textit{zero-shot integration} of visual and tactile information. 
Our framework can leverage any capable visual model as its backbone, allowing it to inherit the generalization capabilities of models trained on large-scale visual datasets. 
During our experiments, we found that visual models often fail under the adverse visual conditions faced during object manipulation~(Fig.~\ref{fig:teaser}).
To counter this, we introduce a feasibility check on the visual estimates based on three physical constraints:
(1) \textit{contact constraint}: the object must make contact with the tactile sensor if a positive contact signal is detected;
(2) \textit{penetration constraint}: at any time, the object cannot penetrate the robot (per its proprioception); and
(3) \textit{kinematic constraint}: the object's motion must be physically feasible.

When visual estimation is deemed infeasible based on these constraints, we refine it through a test-time optimization process using tactile and proprioceptive information.
Specifically, we model the gripper–object interaction as a spring–mass system, where the object is pulled by a virtual ``attractive spring'' towards the tactile sensors according to the tactile signals.
Concurrently, a ``repulsive spring'' repels the object to avoid penetration with the robot.
This optimization process allows for robust, real-time tracking even in dynamic, occluded scenarios.

To demonstrate effectiveness, we apply our framework onto two representative novel object visual estimation models~\cite{wen_foundationpose_2024, labbe_megapose_2022}, conducting extensive experiments on a real-world robotic setup without collecting any tactile dataset for fine-tuning.
Our experiments show that ViTa-Zero outperforms the underlying visual models by a large margin.
Our contributions could be summarized as follows:
\begin{enumerate}
    \item We present the first zero-shot visuotactile novel object pose estimation framework, greatly improving the practicality within the visuotactile domain.
    \item We propose a test-time optimization approach with both efficacy and efficiency using tactile and proprioception to refine the visual estimation results.
    \item We evaluate our approach in real-world robotic manipulation scenarios, demonstrating large performance improvements over the underlying visual models~\cite{wen_foundationpose_2024, labbe_megapose_2022}.
\end{enumerate}

\section{Related Work}
In this section, we survey the representative works in pose estimation from two perspectives: vision-based and visuotactile-based approaches.
We also briefly discuss their relevance and implications for manipulation tasks in robotics.

\subsection{Vision-based Pose Estimation}
Traditionally, pose estimation research has focused on instance-level estimation~\cite{xiang_posecnn_2018, wang_densefusion_2019, park_pix2pose_2019, li_mrc-net_2024}, restricting trained models to specific, known objects and therefore limiting their practicality in real-world applications.
Recent works have expanded the scope to category-level~\cite{wang_normalized_2019, chen_learning_20200, lee_tta-cope_2023} and even to novel object estimation~\cite{wen_foundationpose_2024, labbe_megapose_2022, he_onepose_2022, liu_gen6d_2022, lin_sam-6d_2024}, addressing the challenges associated with broader applications.
Notably, FoundationPose~\cite{wen_foundationpose_2024} achieves state-of-the-art performance in novel object pose estimation, leveraging massively scaled synthetic datasets and photorealistic rendering for effective sim-to-real transfer.
However, contact-rich manipulation tasks often introduce adverse conditions (\eg, occlusion, high dynamicity), which makes visual approaches less effective.
Furthermore, applying such scaling law to tactile sensing is non-trivial due to the non-uniformity of tactile sensors and the wider sim-to-real gap.

Pose estimation has been heavily used in state-based policies, \eg, \cite{openai_solving_2019, andrychowicz_learning_2020, handa_dextreme_2023} perform in-hand object re-orientation task using visual pose estimation.
Their experiments, however, typically depend on elaborate setups, such as cages or multi-view systems, and are generally limited to a single object, restricting their real-world applicability. 
Accordingly, state-based policies are more challenging to deploy in real-life environments compared to visuomotor approaches~\cite{chen_visual_2023, chen_system_2022, qi_general_2023}.

\subsection{Visuotactile-based Pose Estimation}
In contrast, visuotactile-based methods offer a different approach by integrating visual inputs with tactile sensing.
\citet{kuppuswamy_fast_2020} perform tactile-based pose estimation using the Iterative Closest Point (ICP) algorithm with a (vision-based) soft bubble sensor that provides dense geometric data of the contact area.
\citet{dikhale_visuotactile_2022} propose to leverage both RGB-D and tactile information to enhance in-hand object pose estimation.
Later works have explored various techniques, such as applying graph neural networks on taxel signals~\cite{tu_posefusion_2023, rezazadeh_hierarchical_2023}, integrating tactile sensing with proprioception~\cite{rezazadeh_hierarchical_2023}, using explicit shape completion module~\cite{li_vihope_2023}, and hyper-resoluting the sparse taxel signals~\cite{li_hypertaxel_2024}.
While these works have shown great promise, they often face challenges related to datasets and often overfit to a specific type of tactile sensor and gripper.

The work most closely related to ours is by \citet{liu_enhancing_2024}, which also uses tactile signals to refine visual results.
However, their method relies on tracking object velocity by training an ad-hoc model and calculating the optical flow of marker points~\cite{taylor_gelslim_2022} on two vision-based tactile sensors mounted on a parallel gripper. 
This approach presents challenges when applied to non-visual sensors, where optical flow is difficult to obtain.
In contrast, our zero-shot framework leverages physical constraints and imposes fewer conditions on the type of tactile sensor and end-effectors used, making it more versatile for real-world applications.

\section{Problem Definition}
Given a rigid object $\mathcal{O}$, we estimate (in the camera's coordinate frame) its pose ${\mathit{T} = (\mathit{R}, \mathbf{t}) \in \mathbb{SE}(3)}$, comprised of its orientation ${\mathit{R} \in \mathbb{SO}(3)}$ and position ${\mathbf{t} \in \R^3}$.
Following the problem setup of novel object pose estimation~\cite{wen_foundationpose_2024, labbe_megapose_2022}, we assume the 3D mesh $\mathit{M}_\mathcal{O}$ of the object $\mathcal{O}$ is either given or reconstructed~\cite{wen_bundlesdf_2023, muller_instant_2022}.
The object $\mathcal{O}$ is observed using an RGB-D sensor and visible for at least the initial frame.

\textbf{Robot.}
The object $\mathcal{O}$ is manipulated by the robot $\mathcal{R}$ with rigid links and joints. 
We assume the robot kinematic model and the part meshes are given, \eg, through Unified Robot Description Format (URDF), to enable an accurate calculation of the robot's end-effector poses.
The relative transform between the end-effectors and camera frames is obtained using hand-eye calibration~\cite{marchand_visp_2005}.

\textbf{Sensor.}
The robot~$\mathcal{R}$ is equipped with tactile sensors~$\mathcal{S}$, and their installation positions are known.
We assume that the sensors will only make contact with the object $\mathcal{O}$ and no self-contact.
Although we focus on taxel-based sensors in this paper, our framework can extend to other types of sensors, which we will discuss in the next section.

\section{Methodology}
\label{sec: method}

\begin{figure}[t!]
    \vspace{0.15cm}
    \centering
    \includegraphics[width=\columnwidth]{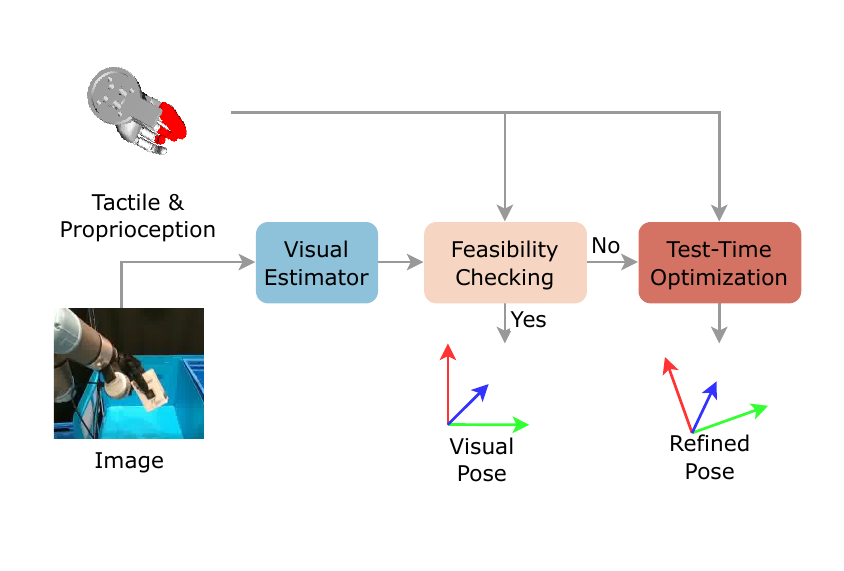}
    \vspace{-3em}
    \caption{
    \textbf{Overview of ViTa-Zero.}
    Red fingers in the robot hand model represent activated fingertip tactile sensors.
    }
    \label{fig: method}
\end{figure}
Our framework consists of three modules: visual estimation, feasibility checking, and tactile refinement.
An overview of our framework is in Fig.~\ref{fig: method}.
Initially, a visual model estimates the pose denoted as $\mathit{T}$.
Then, we assess the feasibility of $\mathit{T}$ using constraints derived from the tactile signals and proprioception.
If $\mathit{T}$ does not meet these constraints, we refine it through our test-time optimization algorithm using tactile and proprioceptive observations, yielding the final pose estimate, denoted as $\mathit{T}^*$.

\subsection{Visual Estimation}
\begin{figure*}[t!]
    \vspace*{0.15cm}
    \centering
    \includegraphics[width=\linewidth]{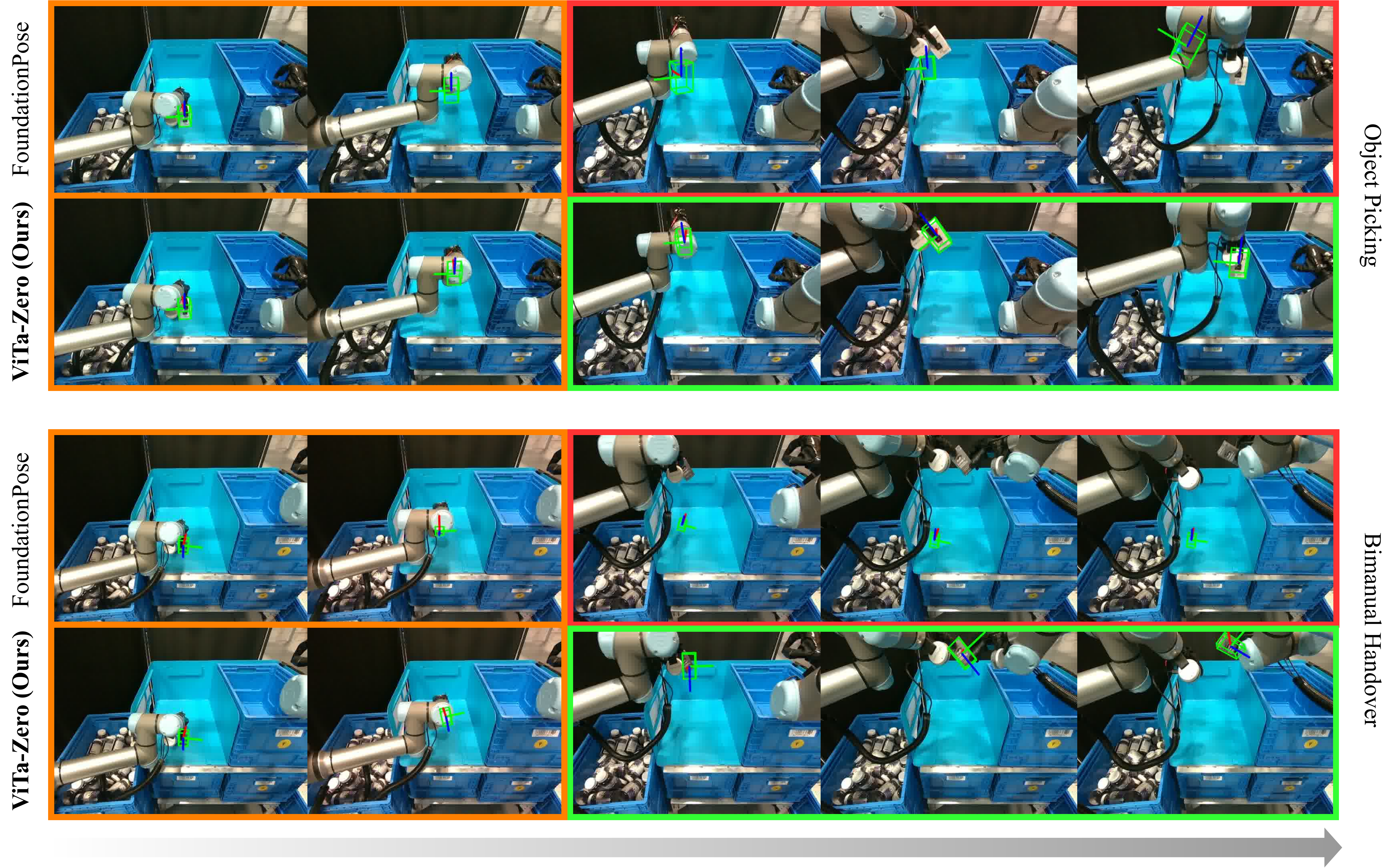}
    \caption{
    \textbf{Qualitative results.}
    We demonstrate the performance during the object picking and bimanual handover tasks with ``camera'' and ``eyedrop'' objects.
    During these manipulation tasks, it is common to encounter scenarios where the object is \textcolor{Orange}{\textbf{highly occluded while moving}}, as illustrated in the figure.
    Visual approaches, like FoundationPose, can \textcolor{Red}{\textbf{lose tracking}} and fail due to the absence of visual information. 
    In contrast, our method utilizes additional tactile and proprioceptive feedback to \textcolor{Green}{\textbf{maintain object tracking}}, ensuring robust performance.
    We note that \textbf{the wrist cameras were not used} in this study.
    }
    \label{fig: qualitative}
    \vspace{-1em}
\end{figure*}

We pass the visual observation into a visual model to obtain the initial pose estimate $\mathit{T}$.
In this paper, we utilized MegaPose~\cite{labbe_megapose_2022} for RGB input and FoundationPose~\cite{wen_foundationpose_2024} for RGB-D input.
We follow the estimate-then-track procedure: we utilize the estimation model for the first frame and the pose tracking model for subsequent frames to estimate the relative pose between consecutive frames.
Our experiments show that the visual models struggle to cope with severe occlusions and dynamic scenes due to the lack of visual cues, leading to significant failures and accumulative errors during manipulation tasks (Fig.~\ref{fig: qualitative}).
Therefore, refining visual results with tactile sensing is crucial for reliable manipulation.

\subsection{Feasibility Checking}

We introduce the geometric representations of each observation used for feasibility checking: tactile signals, the object, and the robot (proprioception).

\textbf{Representation for the tactile signals.}
Prior studies have represented tactile signals as image~\cite{villalonga_tactile_2021, caddeo_collision-aware_2023, guzey_dexterity_2023, yang_binding_2024, dou_tactile-augmented_2024, zhao_transferable_2024} and raw values~\cite{lin_learning_2024, yin_rotating_2023}.
However, these representations do not generalize well across different sensors, \eg image-based methods are ineffective for taxel-based or F/T sensors.
To improve generalization across different tactile sensors, we represent the tactile signal as a point cloud ${\mathit{P}_\mathcal{S}}$ in the camera coordinate system, whose positions could be obtained through forward kinematics.
For taxel-based sensors, such as those used in this paper, point values (in Boolean) are obtained by discretizing sensor outputs with a binary threshold~\cite{yin_rotating_2023, xue_arraybot_2023, li_vihope_2023, dikhale_visuotactile_2022}.
For vision-based sensors~\cite{lambeta_digit_2020, yuan_gelsight_2017, donlon_gelslim_2018, taylor_gelslim_2022}, the point cloud could be obtained through depth estimation~\cite{bauza_tactile_2019, suresh_neuralfeels_2024, kuppuswamy_fast_2020, suresh_midastouch_2023, suresh_shapemap_2022, ambrus_monocular_2021}.

\textbf{Representation for the object.}
Given the object mesh $\mathit{M}_\mathcal{O}$, we obtain the model point cloud $\mathit{P}_\mathcal{O}$ for contact constraint checking by randomly sampling points on the mesh.
$\mathit{P}_\mathcal{O}$ is further downsampled using voxelization to create a voxel grid $\mathit{V}_\mathcal{O}$, allowing faster collision checks.

\textbf{Representation for the robot.}
We use a mesh representation $\mathit{M}_\mathcal{R}$ for the robot $\mathcal{R}$, parameterized by joint angles, instead of directly using joint angles or end-effectors' positions~\cite{rezazadeh_hierarchical_2023, lee_making_2020, lin_learning_2024, yin_rotating_2023} due to the latter's specificity to robot embodiments and less direct physical interpretation. 
This mesh is converted into a downsampled point cloud $\mathit{P}_\mathcal{R}$, using the same downsampling techniques applied to the object mesh to accelerate feasibility checks.

\subsubsection{Contact constraint}
\textit{Each point in the tactile point cloud $\mathit{P}_\mathcal{S}$ must be in contact with some point on the object model point cloud.}
The minimum pairwise Euclidean distance between a point $\mathbf{p}_\mathcal{O}^i$ on the transformed object point cloud $\mathit{P}_\mathcal{O}$ and a point $\mathbf{p}_\mathcal{S}^j$ in the tactile point cloud $\mathit{P}_\mathcal{S}$ must satisfy:
\begin{equation}
    \min_{i,j} \Vert\mathit{T} (\mathbf{p}_\mathcal{O}^i) - \mathbf{p}_\mathcal{S}^j\Vert \leq \theta_\mathrm{c},
\end{equation}
where $\theta_\mathrm{c}$ is the contact distance threshold.

\subsubsection{Penetration constraint}
\textit{No point on the robot model should penetrate the object model and vice versa.}
This constraint is enforced by checking the overlap (intersection) between the transformed object voxel grid $\mathit{T} (\mathit{V}_\mathcal{O})$ and the robot model point cloud $\mathit{P}_\mathcal{R}$.
We use the downsampled robot point cloud $\mathit{P}_\mathcal{R}$ as queries and calculate the proportion of the occupied voxels in the object model:
\begin{equation}
    \frac{\operatorname{card} \big[\mathit{T} (\mathit{V}_\mathcal{O}) \cap \mathit{P}_\mathcal{R}\big]}{\operatorname{card} \big[\mathit{T} (\mathit{V}_\mathcal{O})\big]} \leq \theta_\mathrm{p},
\end{equation}
where $\operatorname{card}[\cdot]$ denotes set cardinality and $\theta_\mathrm{p}$ is the allowable overlapping threshold to compensate for sensor noise and calibration errors.

\subsubsection{Kinematic constraint}
\textit{The object's motion must be physically feasible.}
The following first-order differential constraint on the object's position prevents unrealistic motion, such as pose flipping between frames for symmetrical objects, by ensuring the movement of the object complies with the actual dynamics:
\begin{equation}
    \Vert\mathit{P}_n - \mathit{P}_{n-1}\Vert \leq \theta_\mathrm{d},
\end{equation}
where $\mathit{P}_n \subset \mathit{P}_\mathcal{O}$ represents the contact patch on the object model for $n\textsuperscript{th}$ frame.

\subsection{Test-time Optimization}
If the visual pose estimate $\mathit{T}$ fails to meet any constraints, we refine it using a test-time optimization approach.
We employ a spring–mass model comprising attractive and repulsive springs.
The spring generates elastic potential energy in the form of $\frac{1}{2}kx^2$ (from Hooke's Law), where $x$ is the displacement of the spring.
Unlike the previous works that apply spring–mass models for hand grasp synthesis~\cite{yang_cpf_2021, yang_learning_2024_tpami, brahmbhatt_contactgrasp_2019}, we repurpose it for object pose refinement based on tactile and proprioceptive feedback.
The optimization objective is to \textit{find a relative pose $\mathit{T}_\Delta$ that minimizes the total potential energy from both springs}.

The attractive spring pulls the object towards the tactile sensors having \textit{in-contact} states and reaches its rest length when the object is in perfect contact with all such sensors.
The potential energy of attractive spring is defined as ${E_\mathrm{a} = \frac{1}{2} k_\mathrm{a} \, \underset{i,j}{\min} \, \Vert \mathit{T}_\Delta (\mathbf{p}_\mathcal{O}^i) - \mathbf{p}_\mathcal{S}^j \Vert^2}$.
The repulsive spring prevents penetration between the robot model~$\mathit{P}_\mathcal{R}$ and the object model~$\mathit{P}_\mathcal{O}$.
We define the potential energy as ${E_\mathrm{r} = \frac{1}{2} k_\mathrm{r} \max(0, \gamma)^2}$ where $\gamma$ is the penetration distance defined as:
\begin{equation}
    \gamma = \frac{1}{\operatorname{card}(\mathit{P}_\mathcal{R})} \sum_i (\widetilde{\mathbf{p}}_\mathcal{O}^i - \mathbf{p}_\mathcal{R}^i) \cdot \widetilde{\mathbf{n}}_\mathcal{O}^i.
\end{equation}
Here, $\widetilde{\mathbf{p}}_\mathcal{O}^i = \underset{\mathbf{p} \in \mathit{P}_\mathcal{O}}{\operatorname{argmin}} \, \Vert \mathbf{p} - \mathit{T}_\Delta (\mathbf{p}_\mathcal{R}^i) \Vert$ represents the nearest neighbor of the robot's  refined point $i$; and $\widetilde{\mathbf{n}}_\mathcal{O}^i$ is its unit normal on $\mathit{P}_\mathcal{O}$ pointing outward of the mesh surface.
Hence, $\gamma$ is the mean of signed distances along $\widetilde{\mathbf{n}}_\mathcal{O}^i$ between points $\mathit{P}_\mathcal{R}$ and their nearest neighbors in $\mathit{P}_\mathcal{O}$.

We parameterize the relative pose $\mathit{T}_\Delta$ into relative rotation $\mathit{R}_\Delta$ (expressed in angle–axis form using the RoMa library~\cite{bregier_deep_2021}) and relative translation $\mathbf{t}_\Delta$.
We apply $L_2$ regularization on the parameters to stabilize the optimization.
The refined pose $\mathit{T}^*$ is obtained by:
\begin{equation}
\begin{split}
    \mathit{T}_\Delta &=  \argmin_{\mathit{R}_\Delta, \mathbf{t}_\Delta} (E_\mathrm{a} + E_\mathrm{r} + \lambda \mathcal{L}_2), \\
    \mathit{T}^* &= \mathit{T}_\Delta \circ \mathit{T},
\end{split}
\end{equation}

where $\mathcal{L}_2$ denotes the $L_2$ regularization loss.
In the subsequent frame, we perform pose tracking on the $\mathit{T}^*$.

We implement our algorithm using PyTorch~\cite{paszke_pytorch_2019} and Adam~\cite{kingma_adam_2017} solver.
To facilitate the optimization process, we initialize the $\mathit{T}_\Delta$ using the average translation changes of the activated taxels.
This initialization strategy smoothes the optimization process and improves the overall performance. 

\begin{figure}[t!]
    \vspace{0.15cm}
    \centering
    \includegraphics[width=\columnwidth]{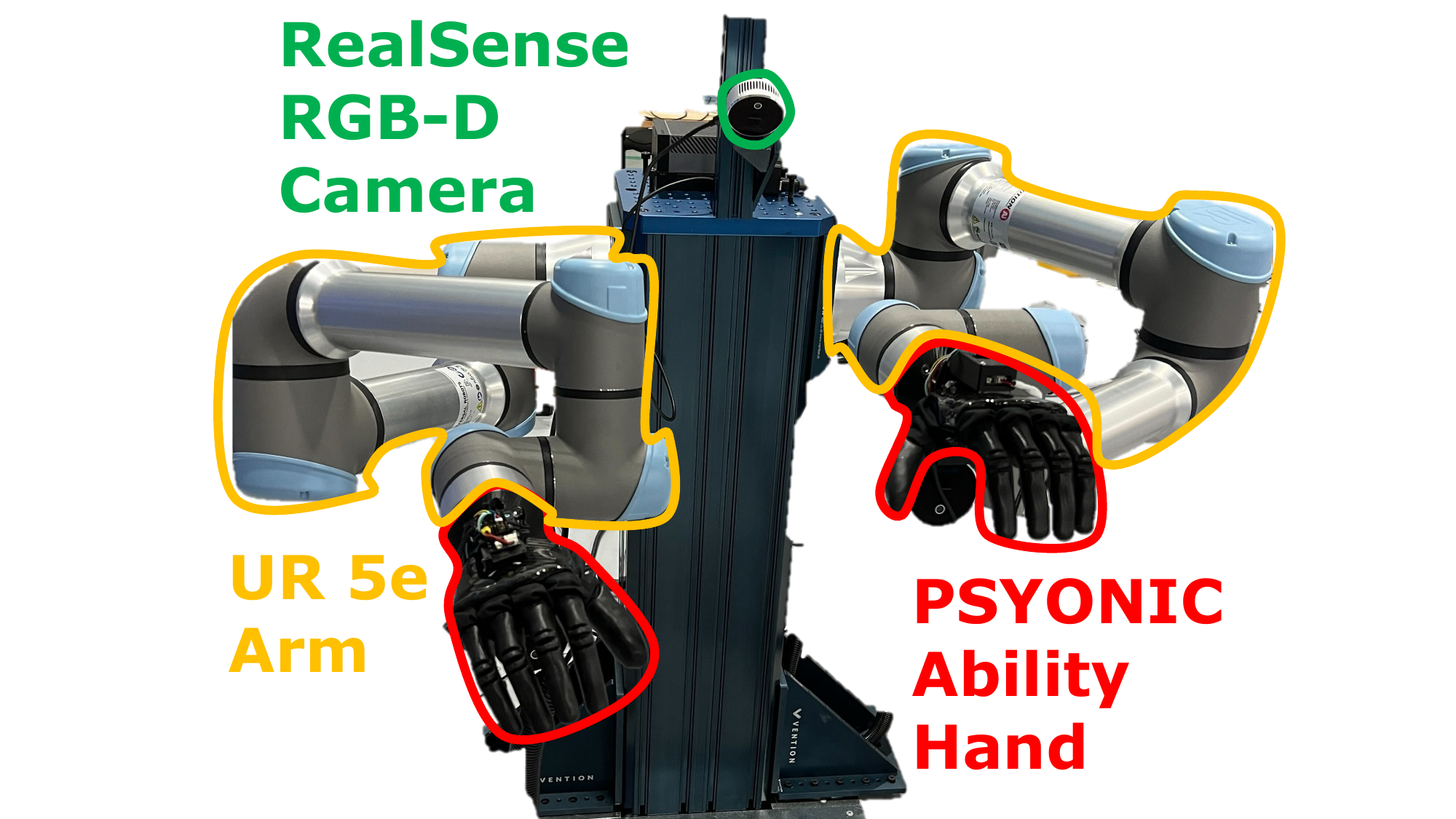}
    \caption{
    \textbf{Our robot platform.}
    Our setup consists of two Universal Robots UR 5e arms and PSYONIC Ability hands.
    }
    \label{fig: robot}
    \vspace{-1em}
\end{figure}

\section{Experiments}

\begin{figure}[t!]
    \vspace{0.15cm}
    \centering
    \includegraphics[width=\columnwidth]{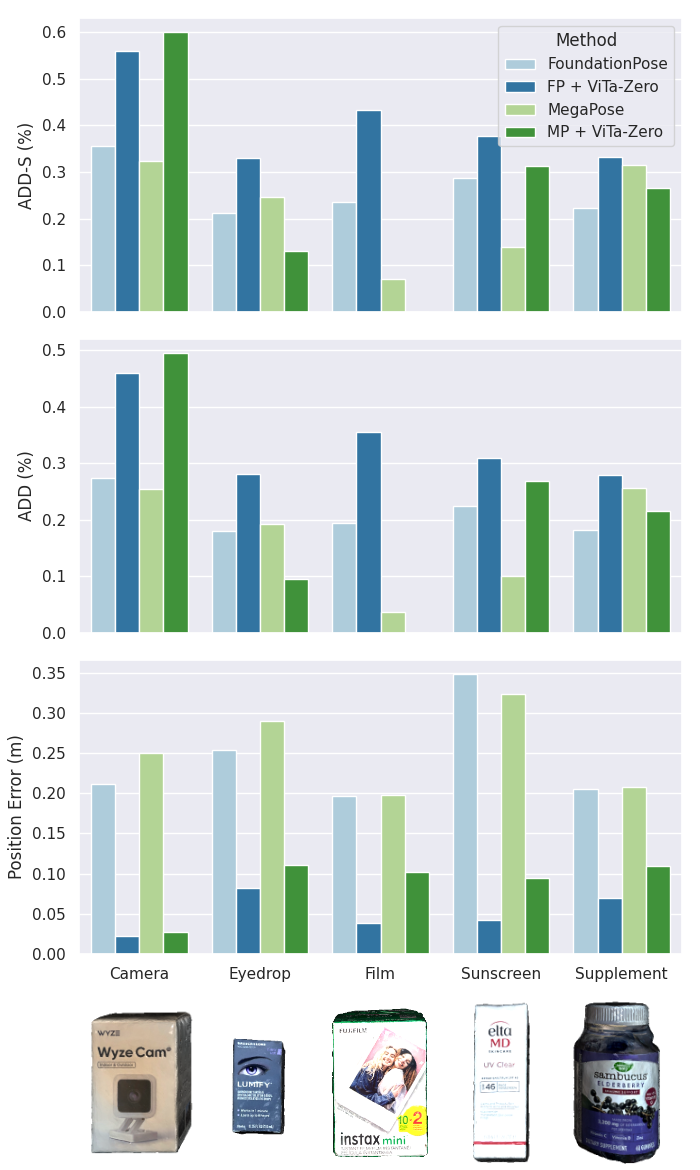}
    \caption{
    \textbf{Comparison with FoundationPose (FP) and MegaPose (MP).}
    The performance is measured by the AUC of ADD-S and ADD (the higher, the better) and position error (PE) (the lower, the better).
    }
    \label{fig: objects}
    \vspace{-1em}
\end{figure}

We evaluate our approach on a real-world robot platform (Fig.~\ref{fig: robot}), consisting of two Universal Robots UR5e robot arms and two PSYONIC Ability hands.
The anthropomorphic hand has five fingers, each equipped with six FSR sensors on the fingertip to provide tactile sensing.
More details regarding this platform are available in \cite{lin_learning_2024}.
Visual sensing is provided by RealSense RGB-D cameras, specifically a combination of the D455 and L515 models.
While our framework is designed to generalize across various embodiments and tactile sensors, this paper focuses on the Ability hand platform.
We test on five daily objects (Fig.~\ref{fig: objects}) purchased on {\href{https://www.amazon.com/}{Amazon.com}}, which could be easily held by the Ability hand and, to our best knowledge, are not part of the Objaverse~\cite{deitke_objaverse_2023} and GSO dataesets~\cite{downs_google_2022}.
Therefore, these five objects are considered novel for the following experiments.

In the feasibility checking, we set the contact threshold $\theta_\mathrm{c}=0.05$, the allowable penetration threshold $\theta_\mathrm{p}=0.008$, and the kinematic threshold $\theta_\mathrm{d}=0.03$.
For the test-time optimization algorithm, the optimizer's learning rate is set to $10^{-3}$, and the number of iterations is empirically tuned to 10.
The loss function parameters are set as follows: attractive energy weight $k_\mathrm{a}=1$, repulsive energy weight $k_\mathrm{r}=1000$, and $L_2$ regularization parameter $\lambda=1000$.

In the following experiments, we compare our method with the state-of-the-art novel object pose estimation models: MegaPose (RGB) and FoundationPose (RGB-D)\footnote{We implement FoundationPose using their released checkpoint at \url{https://github.com/NVlabs/FoundationPose}. Since the released checkpoint was trained on a slightly different dataset, it achieves similar but not the exact performance reported in their paper.}.
We obtain the segmentation mask and bounding box needed for these models using off-the-shelf models~\cite{kirillov_segment_2023, liu_grounding_2024}.

\subsection{Qualitative Results}
Figure~\ref{fig: qualitative} presents qualitative results for object picking and bimanual handover tasks.
We observe that the visual model often loses tracking during manipulation (\eg, grasping) causing catastrophic failures.
In contrast, our approach effectively leverages tactile and proprioception to refine the visual estimation result and maintain object tracking.

\subsection{Quantitative Results}
Due to the lack of public visuotactile object pose estimation datasets\footnote{\cite{dikhale_visuotactile_2022, li_vihope_2023, rezazadeh_hierarchical_2023, li_hypertaxel_2024}'s datasets are proprietary, and \cite{wan_vint-6d_2024, suresh_neuralfeels_2024} are not released by the time we finish this paper.}, we collect a dataset for testing purposes.
We evaluate the performance using position error (PE) and the area under the curve (AUC) of ADD and ADD-S~\cite{li_vihope_2023, dikhale_visuotactile_2022, xiang_posecnn_2018, wen_foundationpose_2024}.
PE is determined as the $L_2$ norm between the ground-truth translation and the estimated translation.

\subsubsection{Comparison with baselines}
\begin{table}[h!]
\caption{Quantitative comparison with FoundationPose (FP) and MegaPose (MP). The best results are bolded.}
 \label{tab: comparison}
\centering
 \begin{tabular}{ l | r r r } 
 \Xhline{4\arrayrulewidth}
 Method &  ADD-S $\uparrow$ & ADD $\uparrow$ & PE $\downarrow$ \\ 
 \hline
FoundationPose & 26.30 & 21.09 & 0.24 \\
\textbf{FP + ViTa-Zero} & \textbf{40.68} & \textbf{33.68} & \textbf{0.05} \\
\hline
MegaPose & 21.90 & 16.88 & 0.25 \\
\textbf{MP + ViTa-Zero} & \textbf{26.19} & \textbf{21.48} & \textbf{0.09} \\
\Xhline{4\arrayrulewidth}
 \end{tabular}
\end{table}
We leverage FoundationPose~\cite{wen_foundationpose_2024} and MegaPose~\cite{labbe_megapose_2022} as our backbones and compare the performance against them (Fig.~\ref{fig: objects} and Tab.~\ref{tab: comparison}).
Since MegaPose lacks depth information, \textit{MegaPose} and \textit{MP + ViTa-Zero} have lower performance than \textit{FoundationPose} and \textit{FP + ViTa-Zero}.
Nonetheless, our framework demonstrates a consistent performance improvement upon its visual backbone, achieving an average increase of 55\% in the AUC of ADD-S and 60\% in ADD, along with an 80\% lower PE compared to FoundationPose.
Despite these significant improvements, the error rate remains relatively high compared to other computer vision benchmarks~\cite{hodan_bop_2024}, highlighting the inherent challenges of pose estimation during manipulation tasks.

\subsubsection{Ablation studies on each loss term}

\begin{table}[h!]
\caption{Ablation studies on loss terms.}
 \label{tab: ablation_loss}
\centering
 \begin{tabular}{ l | r r r } 
 \Xhline{4\arrayrulewidth}
 Ablations &  ADD-S & ADD & PE \\ 
 \hline
Without Attractive Loss & 39.91 & 32.70 & 0.06 \\
Without Penetration Loss & 37.04 & 30.83 & \textbf{0.05} \\
Without $L_2$ Loss & 37.24 & 30.68 & \textbf{0.05} \\
\hline
All Loss & \textbf{40.68} & \textbf{33.68} & \textbf{0.05} \\
\Xhline{4\arrayrulewidth}
 \end{tabular}
\end{table}

We perform ablation studies to assess the impact of each loss term on our approach (Tab.~\ref{tab: ablation_loss}).
When the attractive loss is removed, the refined pose remains static, disregarding tactile feedback.
However, the presence of the penetration loss ensures that the pose is still ``pushed" by the hand model as it moves.
Without the penetration loss, the optimization often leads to solutions in which the object model penetrates the robot model.
Without $L_2$ regularization loss, the optimization process becomes unstable and more challenging to converge.

Overall, our empirical results suggest that excluding any of these three loss terms lead to a decrease in performance, as measured by ADD and ADD-S metrics.

\subsubsection{Ablation studies on refinement algorithm}
\begin{table}[h!]
\caption{Ablation studies on refinement algorithm.}
 \label{tab: ablation_refinement}
\centering
 \begin{tabular}{ l | r r r } 
 \Xhline{4\arrayrulewidth}
 Method &  ADD-S & ADD & PE \\ 
 \hline
ICP & 28.37 & 23.64 & \textbf{0.05} \\
\textbf{Ours} & \textbf{40.68} & \textbf{33.68} & \textbf{0.05} \\
\Xhline{4\arrayrulewidth}
 \end{tabular}
\end{table}
A preliminary version of our test-time optimization algorithm can be viewed as performing point cloud registration between the tactile point cloud and the object surface patches~\cite{kuppuswamy_fast_2020}.
In this study, we investigate the effect of using the ICP algorithm~\cite{besl_method_1992} as our refinement algorithm (Tab.~\ref{tab: ablation_refinement}).
Specifically, we calculate the $\mathit{T}_\Delta$ by registering the object point cloud $\mathit{P}_\mathcal{O}$, serving as the source, with the tactile point cloud $\mathit{P}_\mathcal{S}$, used as the target.
In this paper, we employ the point-to-point ICP estimation method, which minimizes the Euclidean distance between corresponding points.

Our experiments demonstrate that the ICP algorithm effectively refines the visual pose $\mathit{T}$ by aligning it with the nearest tactile points.
However, due to the absence of penetration penalties, the solutions frequently exhibit unintended interpenetration with the robot model.

\subsubsection{Ablation studies on initialization for optimization}
\label{sec: ablation_init}
\begin{table}[h!]
\caption{Ablation studies on optimization initialization.}
 \label{tab: ablation_init}
\centering
 \begin{tabular}{ l | r r r } 
 \Xhline{4\arrayrulewidth}
 Method &  ADD-S & ADD & PE \\ 
 \hline
Without Init & 39.83 & 35.85 & 0.06 \\
With Init & \textbf{40.68} & \textbf{33.68} & \textbf{0.05} \\
\Xhline{4\arrayrulewidth}
 \end{tabular}
\end{table}
In Table~\ref{tab: ablation_init}, we study the impact of removing the initialization for the test-time optimization algorithm.
To recall, we initialize the optimized parameters $\mathit{T}_\Delta$ using the average translation of the activated taxels.
Given the highly non-convex nature of our optimization problem, we find that this initialization strategy leads to more stable optimization and facilitates convergence.

\subsubsection{Running time}
\begin{table}[h!]
\caption{Running time for each component.}
 \label{tab: runtime}
\centering
 \begin{tabular}{ l | r r r } 
 \Xhline{4\arrayrulewidth}
 Module &  Time (s) & Hz \\ 
 \hline
Pose tracking & 0.011 & 90.9 \\
Feasibility checking & 0.006 & 166.7 \\
Test-time optimization & 0.051 & 19.6\\
\Xhline{4\arrayrulewidth}
 \end{tabular}
\end{table}
Given our focus on manipulation applications, we evaluate the running time of each critical component in our framework.
Our platform is equipped with an Intel i9-14900K CPU and an NVIDIA RTX 4090 GPU.
For the pose tracking speed of FoundationPose, we set the refinement iterations as two.
Overall, our framework could reach an online refresh rate of around 20 Hz, given ten iterations during test-time optimization.

\section{Conclusion}
In this paper, we introduced a novel zero-shot visuotactile object 6D pose estimation framework.
Our approach eliminates the need to collect a tactile dataset and demonstrates robust real-world performance on novel objects.
The proposed test-time optimization algorithm leverages a spring–mass model to refine the infeasible visual estimations using tactile and proprioceptive observations.
Experiment results show that our framework could consistently improve the performance of the visual models by a large margin.
Notably, we achieved an average increase of 55\% in the area under the curve (AUC) of ADD-S and 60\% in ADD, along with an 80\% lower position error on FoundationPose.
We also conducted extensive ablation studies to evaluate the impact of various design choices, including different loss functions, refinement algorithms, and optimization initialization techniques.
For future work, we plan to evaluate more object shapes and integrate our pose estimator into manipulation policies to assess its impact on the overall performance of manipulation tasks.

\textbf{Limitations.}
We identify two limitations in our framework:
i) Following the setting of novel object pose estimation, we assume the 3D CAD model of the object is either provided or can be reconstructed from reference views. 
However, this might be challenging for in-the-wild manipulation applications, where active visuotactile object reconstruction techniques~\cite{suresh_neuralfeels_2024} could be leveraged to mitigate this issue.
ii) Our approach relies on the forward kinematics of the robot, which is available in most commercially available grippers.
However, this might be non-trivial when dealing with deformable or compliant grippers~\cite{odhner_stable_2015}.
Recent works~\cite{chen_fully_2022, li_unifying_2024} offer potential solutions to address this challenge.

\footnotesize{
\bibliographystyle{IEEEtranN}
\bibliography{custom, hongyu_zotero}
}

\end{document}